\documentclass[10pt]{article}
\pdfoutput=1
\usepackage{graphicx,graphics}
\usepackage[square,numbers]{natbib}
\DeclareGraphicsExtensions{.pdf,.png,.jpg}
\usepackage{xspace,xcolor}
\usepackage{fullpage}

\usepackage{amsmath}
\usepackage{amsthm}
\usepackage{amsfonts}

\usepackage{algorithm}
\usepackage{algorithmicx}
\usepackage[noend]{algpseudocode}

\usepackage{amsmath,amsfonts}
\usepackage{amssymb,amsopn}
\usepackage{bm} 

\usepackage{amssymb}
\usepackage{amsmath,amsfonts}
\usepackage{amsthm,amsopn}

\usepackage{bm} 

\newcommand{\ProbOpr}[1]{\mathbb{#1}}

\newcommand{\expect}[2]{%
\ifthenelse{\equal{#2}{}}{\ProbOpr{E}_{#1}}
{\ifthenelse{\equal{#1}{}}{\ProbOpr{E}\left[#2\right]}{\ProbOpr{E}_{#1}\left[#2\right]}}} 
\newcommand{\var}[2]{%
\ifthenelse{\equal{#2}{}}{\ProbOpr{VAR}_{#1}}
{\ifthenelse{\equal{#1}{}}{\ProbOpr{VAR}\left[#2\right]}{\ProbOpr{VAR}_{#1}\left[#2\right]}}} 

\newtheorem{thm}{Theorem}

\newcommand{\eat}[1]{}
\usepackage{hyperref}
\usepackage{url}

\theoremstyle{plain}
\newtheorem{lem}[thm]{Lemma}
\newtheorem{prop}[thm]{Proposition}

\usepackage{lineno}

\begin{document}

%

\title{Learning Piece-wise Linear Models \\ from Large Scale Data for Ad Click Prediction}
\author{Kun Gai$^{1}$, Xiaoqiang Zhu$^{1}$, Han Li$^{1}$, Kai Liu$^{2\dagger}$, Zhe Wang$^{3\dagger}$\\[0.5em]
$^1$ Alibaba Inc.\\
 \texttt{jingshi.gk@taobao.com, \{xiaoqiang.zxq, lihan.lh\}@alibaba-inc.com}\\[0.5em]
$^2$ Beijing Particle Inc.   \\\texttt{liukai@yidian-inc.com} \\[0.5em]
$^3$ University of Cambridge. \\ \texttt{zw267@cam.au.uk} \\[0.5em]
$^\dagger$ contribute to this paper while worked at Alibaba\\[0.5em]
}

%
%
%
%

\maketitle

\begin{abstract}
CTR prediction in real-world business is a difficult machine learning problem with large scale nonlinear sparse data.
In this paper, we introduce an industrial strength solution with model named Large Scale Piece-wise Linear Model (LS-PLM).
We formulate the learning problem with  $L_1$ and $L_{2,1}$ regularizers, leading to a non-convex and non-smooth optimization problem.
Then, we propose a novel algorithm to solve it efficiently, based on directional derivatives and quasi-Newton method.
In addition, we design a distributed system which can run on hundreds of machines parallel and provides us with the industrial scalability.
LS-PLM model can capture nonlinear patterns from massive sparse data, saving us from heavy feature engineering jobs.
Since 2012, LS-PLM has become the main CTR prediction model in Alibaba's online display advertising system, serving hundreds of millions users every day.
\end{abstract}


\section{Introduction}

\label{intro}

Click-through rate (CTR) prediction is a core problem in the multi-billion dollar online advertising industry. To improve the accuracy of CTR prediction, more and more data are involved, making CTR prediction a large scale learning problem, with massive samples and high dimension features.

Traditional solution is to apply a linear logistic regression (LR) model, trained in a parallel manner (Brendan et al. 2013, Andrew \& Gao 2007).
LR model with $L_{1}$ regularization can generate sparse solution, making it fast for online prediction.
Unfortunately, CTR prediction problem is a highly nonlinear problem. In particular, user-click generation involves many complex factors,
like ad quality, context information, user interests, and complex interactions of these factors.
To help LR model catch the nonlinearity, feature engineering technique is explored, which is both time and humanity consuming.

Another direction, is to capture the nonlinearity with well-designed models.
Facebook (He et al. 2014) uses a hybrid model which combines decision trees with logistic regression.
Decision tree plays a nonlinear feature transformation role, whose output is fed to LR model.
However, tree-based method is not suitable for very sparse and high dimensional data (Safavian S. R. \& Landgrebe D. 1990).
(Rendle S. 2010) introduces Factorization Machines(FM), which involves interactions among features using 2nd order functions (or using other given-number-order functions). However, FM can not fit all general nonlinear patterns in data (like other higher order patterns).

In this paper, we present a piece-wise linear model and its training algorithm for large scale data.
We name it Large Scale Piecewise Linear Model (LS-PLM).
LS-PLM follows the divide-and-conquer strategy, that is,
first divides the feature space into several local regions, then fits a linear model in each region,
resulting in the output with combinations of weighted linear predictions.
Note that, these two steps are learned simultaneously in a supervised manner, aiming to minimize the prediction loss.
LS-PLM is superior for web-scale data mining in three aspects:
\begin{itemize}
  \item \textbf{Nonlinearity}. With enough divided regions, LS-PLM can fit any complex nonlinear function.
  \item \textbf{Scalability}.  Similar to LR model, LS-PLM is scalable both to massive samples and high dimensional features. We design a distributed system which can train the model on hundreds of machines parallel. In our online product systems, dozens of LS-PLM models with tens of million parameters are trained and deployed everyday.
  \item \textbf{Sparsity}. As pointed in (Brendan et al. 2013), model sparsity is a practical issue for online serving in industrial setting.
      We show LS-PLM with $L_{1}$ and $L_{2,1}$ regularizer  can achieve good sparsity.
\end{itemize}

The learning of LS-PLM with sparsity regularizer can be transformed to be a non-convex and non-differential optimization problem, which is difficult to be solved. We propose an efficient optimization method for such problems, based on directional derivatives and quasi-Newton method. Due to the ability of capturing nonlinear patterns and scalability to massive data, LS-PLMs have become main CTR prediction models in the online display advertising system in alibaba, serving hundreds of millions users since 2012 early. It is also applied in recommendation systems, search engines and other product systems.

The paper is structured as follows.
In Section \ref{method}, we present LS-PLM model in detail, including formulation, regularization and optimization issues.
In Section \ref{implementation} we introduce our parallel implementation structure.
in Section \ref{exp}, we evaluate the model carefully and demonstrate the advantage of LS-PLM compared with LR.
Finally in Section \ref{conclu}, we close with conclusions.

\begin{figure*}[htbp]
    \vspace{-5pt}
    \centering{\includegraphics[width=0.9\textwidth]{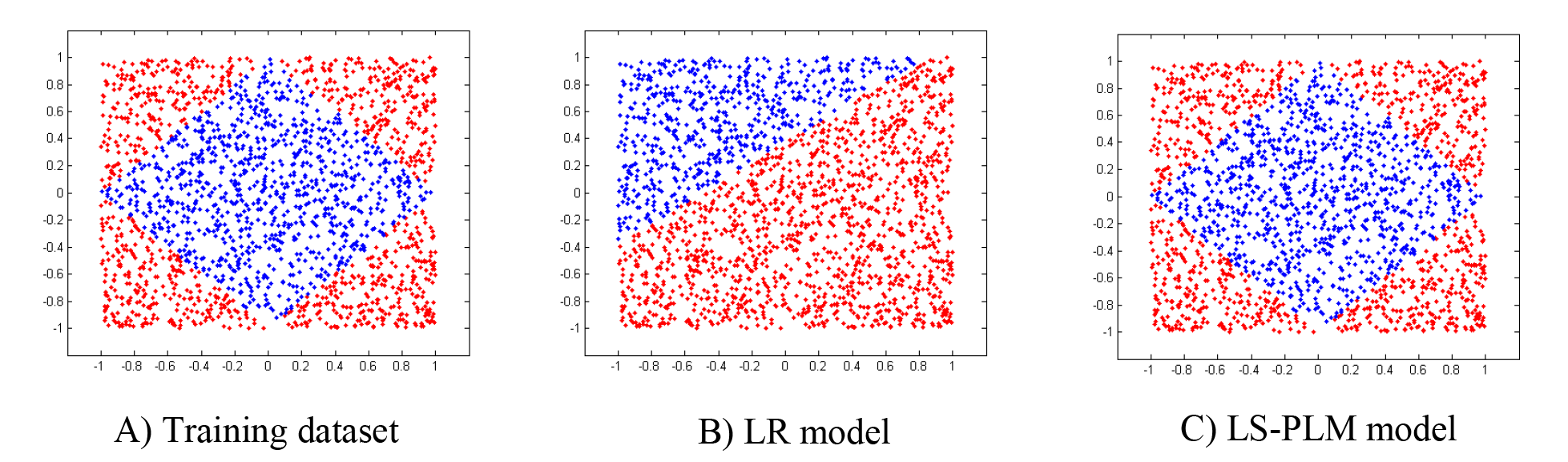}}
    \caption{A demo illustration of LS-PLM model. Figure A is the demo dataset. It is a binary classification problem, with red points belong to positive class and blue points belong to negative class.  Figure B shows the classification result using LR model. Figure C shows the classification result using LS-PLM model. It's clear that LS-PLM can capture the nonlinear distribution of data. }
    \label{demo}
    \vspace{-0pt}
\end{figure*}

\section{Method}
\label{method}
We focus on the large scale CTR prediction application. It is a binary classification problems, with dataset $\{x_t, y_t \}|_{t=1}^n$. $y_t \in \{0,1\}$ and $x_t \in R^d$ is usually high dimensional and sparse.

\subsection{Formulation}
To model the nonlinearity of massive scale data, we employ a divide-and-conquer strategy, similar with (Jordan \& Jacobs 1994). We divide the whole feature space into some local regions. For each region we employ an individual generalized linear-classification model. In this way, we tackle the nonlinearity with a piece-wise linear model. We give our model as follows:
\begin{equation}\label{mlrp2}
p(y=1|x) =g\Big(\sum_{j=1}^m \sigma(u_j^Tx)\eta(w_j^Tx)\Big)
\end{equation}
Here $\Theta=\{u_1,\cdots,u_m,w_1,\cdots,w_m\}\in \mathbb{R}^{d\times 2m}$  denote the model parameters. $\{u_1,\cdots,u_m\}$ is the parameters for dividing function $\sigma(\cdot)$, and $\{w_1,\cdots,w_m\}$ for fitting function $\eta(\cdot)$. Given instance $x$, our predicating model $p(y|x)$ consists of two parts: the first part $\sigma(u_j^Tx)$ divides feature space into $m$ (hyper-parameter) different regions, the second part  $\eta(w_j^Tx)$ gives prediction in each region. Function $g(\cdot)$ ensures our model to satisfy the definition of probability function.

\textbf{Special Case.}
Take softmax ( Kivinen \& Warmuth 1998) as dividing function $\sigma(x)$  and sigmoid (Hilbe 2009) as fitting function $\eta(x)$ and $g(x)=x$, we get a specific formulation:

\begin{equation}\label{mlr-spc}
p(y=1|x) =\sum_{i=1}^m \frac{\exp(u_i^Tx)}{\sum_{j=1}^m \exp(u_j^Tx)} \cdot \frac{1}{1+\exp{(-w_i^Tx)}}
\end{equation}

In this case, our mixture model can be seen as a FOE model (Jordan \& Jacobs 1994, Wang \& Puterman 1998) as follows:
\begin{equation} \label{mlrp}
   p(y=1|x)=\sum_{i=1}^{m} p(z=i | x)p(y|z=i , x)
\end{equation}

Eq.(\ref{mlr-spc}) is the most common used formulation in our real applications.
In the reminder of the paper, without special declaration,
we take Eq.(\ref{mlr-spc}) as our prediction model.
Figure \ref{demo} illustrates the model compared with LR in a demo dataset, which shows clearly LS-PLM can capture the nonlinear pattern of data.

The objective function of LS-PLM model is formalized as Eq.(\ref{mlrobj}):
\begin{equation}   \label{mlrobj}
arg\min{}_\Theta  f(\Theta) = \text{loss}(\Theta) + \lambda \|\Theta\|_{2,1}  +\beta \|\Theta\|_1
\end{equation}
\begin{equation} \label{mlrloss}
\text{loss}(\Theta) \!=\!
     \!- \sum_{t=1}^n  \Big[ y_t\log {(p(y_t\!=\!1| x_t,\Theta))}
      + (1-y_t)\log (p(y_t \! = \!0|x_t,\Theta)) \Big]
\end{equation}

Here $\text{loss}(\Theta)$ defined in  Eq.(\ref{mlrloss})  is the neg-likelihood loss function and  $\|\Theta_{2,1}\|$ and $\|\Theta_{1}\|$ are two regularization terms providing different properties.
First, $L_{2,1}$ regularization ($\|\Theta\|_{2,1} = \sum_{i=1}^{d} \sqrt{\sum_{j=1}^{2m}\theta_{ij}^2}$) is employed for feature selection. As in our model, each dimension of feature is associated with $2m$ parameters.
$L_{2,1}$ regularization are expected to push all the $2m$ parameters of one dimension of feature to be zero, that is, to suppress those less important features.
Second, $L_1$ regularization ($\|\Theta\|_1 = \sum_{ij}|\theta_{ij}|$) is employed for sparsity.
Except with the feature selection property, $L_1$ regularization can also force those parameters of left features to be zero as much as possible, which can help improve the interpretation ability  as well as generalization  performance of the model.

However, both $L_1$ norm and $L_{2,1}$ norm are non-smooth functions.
This causes the objective function of Eq.(\ref{mlrobj}) to be non-convex and non-smooth, making it difficult to employ those traditional gradient-descent optimization methods (Andrew \& Gao 2007,  Zhang 2004, Bertsekas 2003) or EM method (Wang \& Puterman 1998).

Note that, while (Wang \& Puterman 1998) gives the same mixture model formulation as Eq.(\ref{mlrp}), our model is more general for employing different kinds of prediction functions.
Besides, we propose a different objective function for large scale industry data, taking the feature sparsity into consideration explicitly.
This is crucial for real-world applications, as prediction speed and memory usage are two key indicators for online model serving.
Furthermore, we give a more efficient optimization method to solve the large-scale non-convex problem, which is described in the following section.

\subsection{Optimization}

Before we present our optimization method, we establish some notations and definitions that will be used in the reminder of the paper.
Let $\partial_{ij}^+ f(\Theta)$ denote the right partial derivative of $f$ at $\Theta$ with respect to $\Theta_{ij}$:
\begin{equation}
  \partial_{ij}^+ f(\Theta) = \lim_{\alpha \downarrow 0} \frac{f(\Theta + \alpha e_{ij}) - f(\Theta)}{\alpha}
\end{equation}
where $e_{ij}$ is the ${ij}^{\text{th}}$ standard basis vector. The directional derivative of $f$ as $\Theta$ in direction $d$ is denoted as $f'(\Theta;d)$, which is defined as:
\begin{equation}\label{dirder}
   f'(\Theta;d)=\lim_{\alpha \downarrow 0} \frac{f(\Theta+\alpha d) - f(\Theta)}{\alpha}
\end{equation}
A vector $d$ is regarded as a descent direction if $f'(\Theta;d) < 0$.   $\text{sign}(\cdot)$ is the sign function takes value $\{-1,0,1\}$. The projection function 
\begin{equation}\label{proj}
\pi_{ij}(\Theta;\Omega) = \begin{cases}
\Theta_{ij} & ,\quad \text{sign}(\Theta_{ij}) = \text{sign}(\Omega_{ij}) \\
            0&,\quad \text{otherwise}
\end{cases}
\end{equation}
means projecting $\Theta$ onto the orthant defined by $\Omega$.

\subsubsection{\textbf{Choose descent direction}}
As discussed above, our objective function for large scale CTR prediction problem is both non-convex and non-smooth. Here we propose a general and efficient optimization method to solve such kind of non-convex problems. Since the negative-gradients of our objective function do not exists for all $\Theta$, we take the direction $d$ which minimizes the directional derivative of $f$ with $\Theta$ as a replace.  The directional derivative $f'(\Theta;d)$ exists for any $\Theta$ and direction $d$, whic is declared as Lemma \ref{lem1}.
\begin{lem}
  When an objective function $f(\Theta)$ is composed by a smooth loss function with $L_1$ and $L_{2,1}$ norm, for example the objective function given in Eq. (\ref{mlrobj}), the directional derivative $f'(\Theta;d)$ exists for any $\Theta$ and direction $d$. \label{lem1}	
\end{lem}

We leave the proof in Appendix \ref{app_a}.  Since the directional derivative $f'(\Theta;d)$ always exists, we choose the direction as the descent direction which minimizes the directional derivative $f'(\Theta;d)$  when the negative gradient of $f(\Theta)$ does not exist.  The following proposition \ref{prop1} explicitly gives the direction.

 \begin{prop} \label{prop1}
     Given a smooth loss function $\text{loss}(\Theta) $ and an objective function $ f(\Theta) = \text{loss}(\Theta) + \lambda \|\Theta\|_{2,1}  +\beta \|\Theta\|_1$, the bounded direction $d$ which minimizes the directional derivative $f'(\Theta;d)$ is denoted as follows:

\begin{equation}\label{mlrdir}
d_{ij} =
\begin{cases}
      s -\beta\text{sign}(\Theta_{ij}), & \Theta_{ij} \neq 0 \\
      \max\{|s| - \beta,0 \}\text{sign}(s),
      & \Theta_{ij}=0, \|\Theta_{i\cdot}\|_{2,1} \neq 0 \\
      \frac{\max\{\|v\|_{2,1} - \lambda, 0 \}}{\|v\|_{2,1}} v ,& \|\Theta_{i\cdot}\|_{2,1} = 0,
\end{cases}
\end{equation}
 where $ s = -\nabla loss(\Theta)_{ij}- \lambda\frac{\Theta_{ij}}{\|\Theta_{i\cdot}\|_{2,1}} $ and
 $v = \max\{|-\nabla \text{loss}(\Theta)_{ij}| - \beta,0\}\text{sign}(-\nabla \text{loss}(\Theta)_{ij}) $.

\end{prop}

More details about the proof can be found in Appendix \ref{app_b}.
According to the proof, we can see that the negative pseudo-gradient defined in Gao's work (Andrew \& Gao 2007) is a special case of our descent direction.
Our proposed method is more general to find the descent direction for those non-smooth and non-convex objective functions.

Based on the direction $d^k$ in Eq.(\ref{mlrdir}), we update the model parameters along a descent direction calculated by limited-memory quasi-newton method (LBFGS) (Wang \& Puterman 1998), which approximates the inverse Hessian matrix of Eq.(\ref{mlrobj}) on the given orthant.
Motivated by the OWLQN method (Andrew \& Gao 2007), we also restrict the signs of model parameters not changing in each iteration.
Given the chosen direction $d^k$ and the old $\Theta^{(k)}$, we constrain the orthant of current iteration as follows:
\begin{equation} \label{optorth}
\xi_{ij}^{(k)} = \begin{cases}
\text{sign}(\Theta_{ij}^{(k)}),\quad \Theta_{ij}^{(k)} \neq 0 \\
\text{sign}(d^{(k)}_{ij}),\quad \Theta_{ij}^{(k)} = 0
\end{cases}.
\end{equation}
When $\Theta_{ij}^{(k)} \neq 0$, the new $\Theta_{ij}$ would not change sign in current iteration. When $\Theta_{ij}^{(k)} = 0$, we choose the orthant decided by the selected direction $d^{(k)}_{ij}$ for the new $\Theta_{ij}^{(k)}$.

\begin{algorithm}[t]
\caption{ Optimize problem Eq.(\ref{mlrobj})}
\textbf{Input}:Choose initial point $\Theta^{(0)}$ \\
$S\leftarrow \{\},  Y \leftarrow \{\}$\\
for $k = 0$ to \textbf{MaxIters} do
\begin{enumerate}
  \item Compute  $d^{(k)}$ with Eq. (\ref{mlrdir}).
  \item Compute $p_k$ with Eq. (\ref{optdir}) using $S$ and $Y$.
  \item Find $\Theta^{(k+1)}$ with constrained line search (\ref{optline}).
  \item If termination condition satisfied then \\
        $\text{stop and return }\Theta^{(k+1)}$\\
        End if
  \item Update $S$ with $ s^{(k)} = \Theta^{(k)} - \Theta^{(k-1)}$
  \item Update $Y$ with $ y^{(k)} =  -d^{(k)} - (-d^{(k-1)} )$
\end{enumerate}
\textbf{end for}
\end{algorithm}

%
%

\subsubsection{\textbf{Update direction constraint and line search}}

Given the descent direction $d^k$, we approximate the inverse-Hessian matrix $H_k$ using LBFGS method with a sequence of ${y}^{(k)}, {s}^{(k)}$.
Then the final update direction is $H_kd^{(k)}$. Here we give two tricks to adjust the update direction. First, we constrain the update direction in the orthant with respect to $d^{(k)}$. Second, as our objective function is non-convex, we cannot guarantee that $H_k$ is positive-definite. We use ${y^{(k)T}}s^{(k)}>0$ as a condition to ensure $H_k$ is a positive-definite matrix. If $ y^{(k)T}s^{(k)}\leq 0$, we switch to $d^{(k)}$ as the update direction. The final update direction $p_k$ is defined as follows:
\begin{equation}\label{optdir}
  p_k = \begin{cases}
     \pi(   H_kd^{(k)};   d^{(k)}   ),\quad y^{(k)T}s^{(k)} > 0 \\
    d^{(k)},\quad \text{otherwise}
\end{cases}
\end{equation}

Given the update direction, we use backtracking line search to find the proper step size $\alpha$. Same as OWLQN, we project the new $\Theta^{(k+1)}$ onto the given orthant decided by the Eq. (\ref{optorth}).
\begin{equation} \label{optline}
\Theta^{(k+1)} = \pi (\Theta^{(k)} + \alpha p_k; \xi^{(k)})
\end{equation}

\subsection{Algorithm}
A pseudo-code description of optimization is given in Algorithm 1.
In fact, only a few steps of the standard LBFGS algorithm need to change.
These modifications are:
\begin{enumerate}
\item	The direction $d^{(k)}$ which minimizes the direction derivative of the non-convex objective is used in replace of negative gradient.
\item	The update direction is constrained to the given orthant defined by the chosen direction $d^{(k)}$.
        Switch to $d^{(k)}$ when the $H_k$ is not positive-definite.
\item	During the line search, each search point is projected onto the orthant of the previous point.
\end{enumerate}

\section{Implementation}
In this section, we first provide our parallel implementation of LS-PLM model for large-scale data, then introduce an important trick which helps to accelerate the training procedue greatly.
\label{implementation}

\begin{figure*}[htbp]
\vspace{-5pt}
\centering{\includegraphics[width=1.0\textwidth]{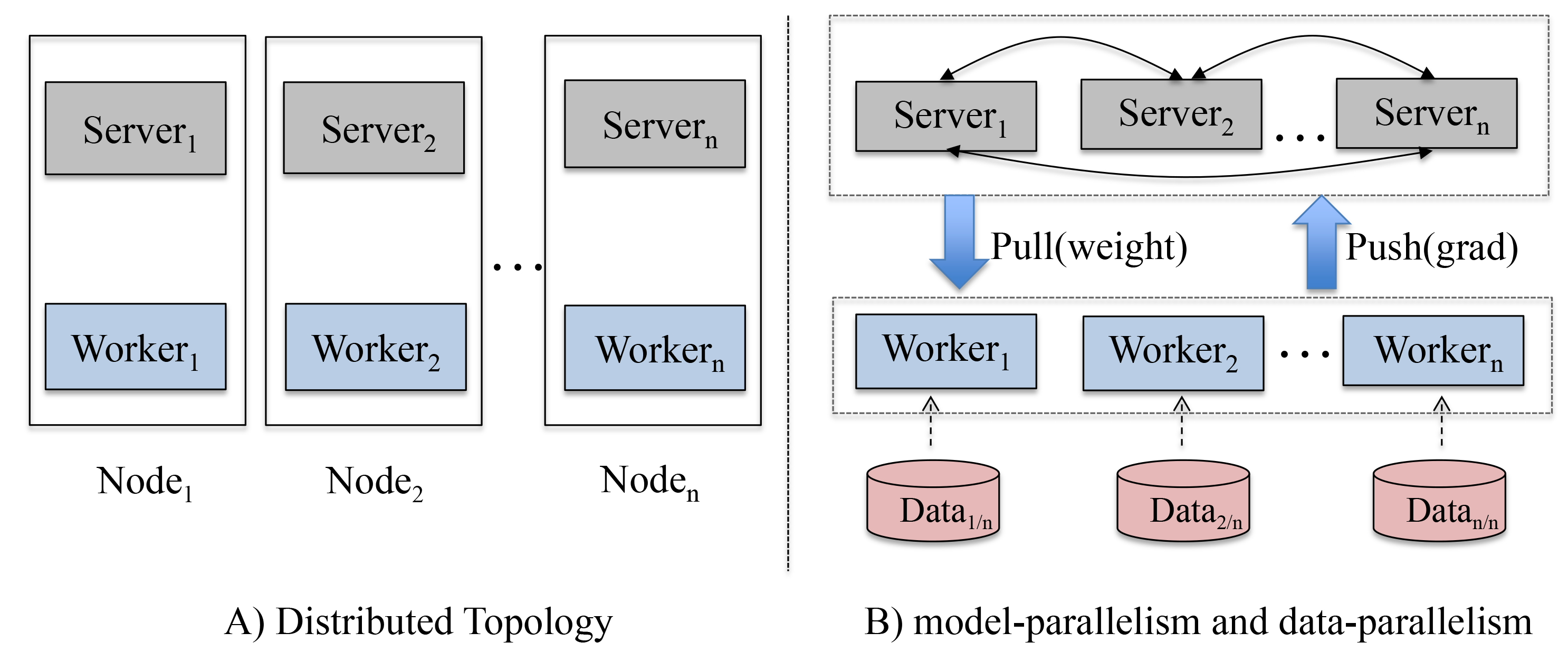}}
\caption{The architecture of parallel implementation.
    Figure A illustrates the physical distributed topology. It's a variant of parameter server, where each computation node runs with both a server and a worker, aiming to maximize the utility of computation power as well as memory usuage. Figure B illustrates the parameter server structure in model-parallelism  and data-parallelism manner. }
\vspace{-0pt}
\label{xone}
\end{figure*}

\subsection{Parallel implementation}

To apply Algorithm 1 in large-scale settings, we implement it with a distributed learning framework, as illustrated in Figure \ref{xone}.
It's a variant of parameter server.
In our implementation, each computation node runs with both a server node and a worker node, aiming to
\begin{itemize}
\item Maximize the utility of CPU's computation power.
      In traditional parameter server setting, server nodes work as a distributed KV storer with interfaces of push and pull operations, which are low computation costly. Running with worker nodes can make full use of the computation power.
\item Maximize the utility of memory. 	
      Machines today usually have big memory, for example 128GB.
      Running on the same computation node, server node and worker node can share and utilize the big memory better.
\end{itemize}

In brief, there are two roles in the framework.
The first role is the work node. Each node stores a part of training data and a local model, which only saves the model parameters used for the local training data.
The second role is the server node. Each node stores a part of global model, which is mutually-exclusive.
In each iteration, all of the worker nodes first calculate the loss and the descent direction with local model and local data in parallel(data parallelism).
Then server nodes aggregate the loss and the direction $d^{(k)}$
as well as the corresponding entries of $\Theta$ needed to calculate the revised gradient (model parallelism).
After finishing calculating the steepest descent direction in Step 1,
workers synchronize the corresponding entries of $\Theta$, and then,
perform Step 2--6 locally.

\subsection{Common Feature Trick}
\begin{figure}[htbp]
    \vspace{-5pt}
    \centering{\includegraphics[width=0.4\textwidth]{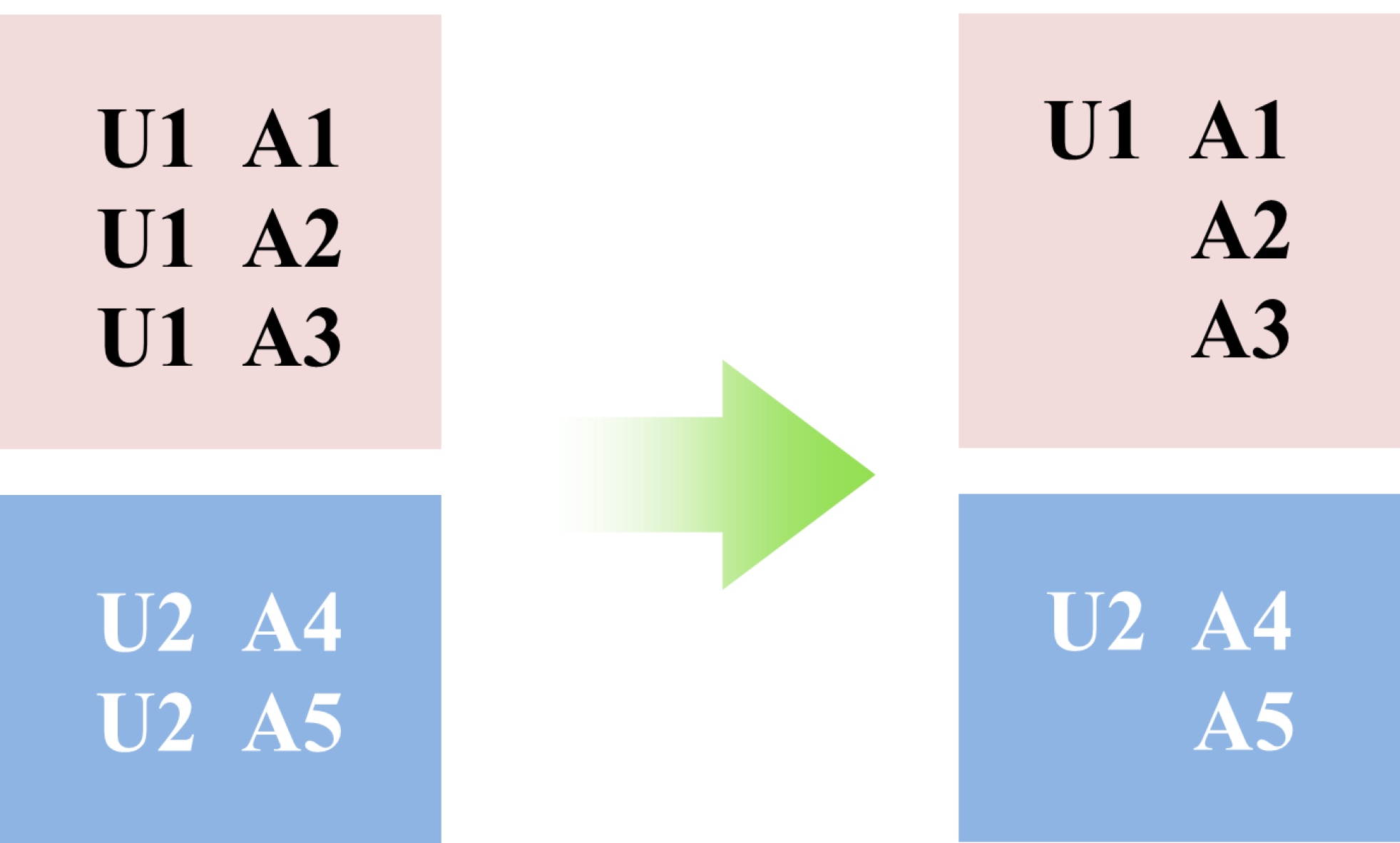}}
    \caption{Common feature pattern in display advertising. Usually in each page view, a user will see several different ads at the same time. In this situation, user features can be shared across these samples.}
    \vspace{-0pt}
    \label{cf_trick}
\end{figure}

In addition to the general-purpose parallel implementation, we also optimized the implementation in online advertising context.
Training samples in CTR prediction tasks usually have similar common feature pattern.
Take display advertising as an example, as illustrated in Figure \ref{cf_trick}, during each page view, a user will see several different ads at the same time.  For example, user \textbf{U1} in Figure \ref{cf_trick} sees three ads in one visit session, thus generates three samples.  In this situation, features of user \textbf{U1}  can be shared across these three samples. These features include user profiles (sex, age, etc.) and user behavior histories during visits of Alibaba¡¯s e-commerce websites, for example, his/her shopping item IDs, preferred brands or favorite shop IDs.

Remember the model defined in Eq. \ref{mlr-spc}, most computation cost focus on $\mu_i^Tx$ and $w_i^Tx$.
By employing the common feature trick, we can split the calculation into common and non-common parts and rewrite them as follows:
\begin{align}
	\mu_i^Tx = \mu_{i,c}^Tx_{c} + \mu_{i,nc}^Tx_{nc} \\ \nonumber
	w_i^Tx = w_{i,c}^Tx_{c} + w_{i,nc}^Tx_{nc}	
\end{align}
Hence, for the common feature part, we need just calculate once and then index the result, which will be utilized by the following samples.

Specifically, we optimize the parallel implementation with common features trick in the following three aspects:
\begin{itemize}
  \item Group training samples with common features and make sure these samples are stored in the same worker.
  \item Save memory by storing common features shared by multiple samples only once.
  \item Speed up iteration by updating loss and gradient w.r.t. common features only once.
\end{itemize}
Due to the common feature pattern of our production data,
employing the common feature trick improves the performance of training procedure greatly, which will be shown in the following section \ref{exp_cf}.

\section{Experiments}
\label{exp}
In this session, we evaluate the performance of LS-PLM.
Our datasets are generated from Alibaba's mobile display advertising product system.
As shown in Table \ref{datasets}, we collect seven datasets in continuous sequential periods, aiming to evaluate the consist performance of the proposed model, which is important for online product serving.
In each dataset, training/validation/testing samples are disjointly collected from different days, with proportion about 7:1:1.
AUC (Fawcett 2006) metric is used to evaluate the model performance.

%
%
%

\begin{table*}
  \centering
  \caption{Alibaba's mobile display advertising CTR prediction datasets}
  \vspace{-5pt}
  \begin{tabular}{ccc} \\
    \hline
    Dataset   & \#features     &     \#samples (training/validation/testing)  \\
    \hline
    1              & $3.04\times 10^6$   & $1.34/0.25/0.26\times 10^9$  \\
    2              & $3.27\times 10^6$   & $1.44/0.26/0.26\times 10^9$   \\
    3              & $3.49\times 10^6$   & $1.56/0.26/0.25\times 10^9$   \\
    4              & $3.67\times 10^6$   & $1.62/0.25/0.26\times 10^9$   \\
    5              & $3.82\times 10^6$   & $1.69/0.26/0.26\times 10^9$   \\
    6              & $3.95\times 10^6$   & $1.74/0.26/0.26\times 10^9$   \\
    7              & $4.07\times 10^6$   & $1.78/0.26/0.26\times 10^9$   \\
    \hline
  \end{tabular}
  \vspace{-5pt}
  \label{datasets}
\end{table*}

\subsection{Effectiveness of division number}
\begin{figure}[htbp]
    \vspace{-5pt}
    \centering{\includegraphics[width=0.5\textwidth]{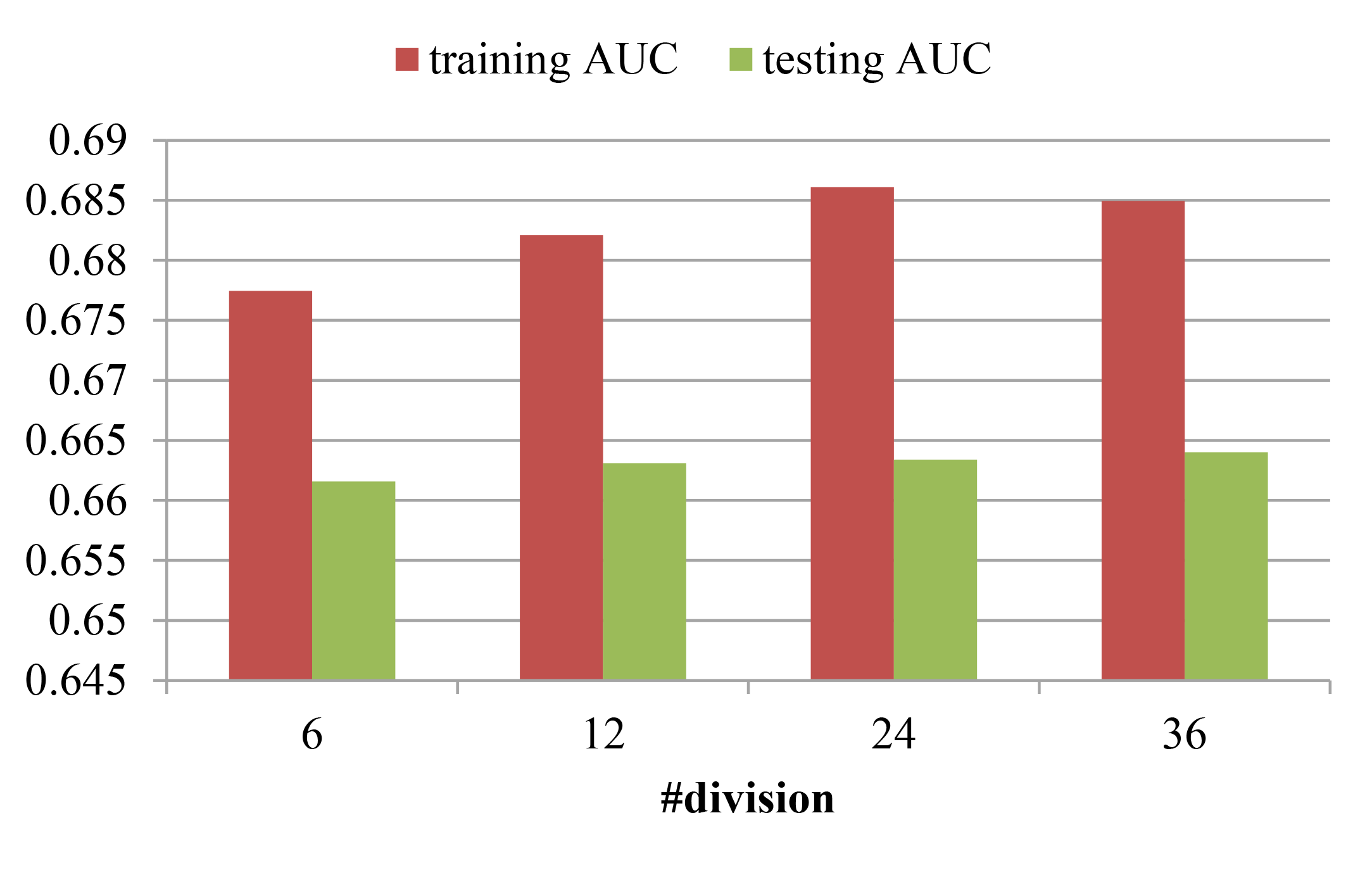}}
    \caption{Model performance with different divisions. }
    \vspace{-0pt}
    \label{exp_m}
\end{figure}

LS-PLM is a piece-wise linear model, with division number \textbf{m} controlling the model capacity.
We evaluate the division effectiveness on model's performance. It is executed on dataset 1 and results are shown in Figure \ref{exp_m}.

Generally speaking, larger $m$ means more parameters and thus leads to larger model capacity. But the training cost will also increase, both time and memory.
Hence, in real applications we have to balance the model performance with the training cost.

Figure \ref{exp_m} shows the training and testing AUC with different division number m.
We try $m=6,12,24,36$, the testing AUC for $m=12$ is markedly better than $m=6$, and improvement for $m=24,36$ is relatively gentle.
Thus in all the following experiments , the parameter $m$ is set as 12 for LS-PLM model.

\subsection{Effectiveness of regularization}

As stated in Session \ref{method}, in order to keep our model simpler and more generalized,
we prefer to constrain the model parameters sparse by both $L_1$ and $L_{2,1}$ norms.
Here we evaluate the strength of both the regularization terms.


\begin{table}
  \centering
  \caption{Regularization effects on model sparsity and performance  }
  \vspace{-5pt}
  \begin{tabular}{cccccc} \\
    \hline
    $\beta/\lambda(L_1/L_{2,1})$ &  \#features  &  \#non-zero parameters & testing auc  \\
    \hline
    0/0          & $3.04 \times 10^6$       & $7.30 \times 10^7$         &  0.6489    \\
    \hline
    0/1          & $5.68 \times 10^5$        & $6.64 \times 10^6$        &  0.6570    \\
    \hline
    1/0          & $3.87 \times 10^5$       & $1.33 \times 10^6$         &  0.6617    \\
    \hline
    1/1         & $2.55 \times 10^5$        & $1.15 \times 10^6$        &  0.6629    \\
    \hline
  \end{tabular}
  \vspace{-5pt}
  \label{regucomp}
\end{table}

Table \ref{regucomp} gives the results.
As expected, both $L_1$ and $L_{2,1}$ norms can push our model to be sparse.
Model trained with $L_{2,1}$ norm has only 9.4\% non-zero parameters left and 18.7\% features are kept back.
While in $L_1$ norm case, there are only 1.9\% non-zero parameters left.
Combining them together, we get the sparsest result.
Meanwhile, models trained with different norm get different AUC performance.
Again adding the two norms together the model reaches the best AUC performance.

In this experiment, the hyper-parameter $m$ is set to be 12. Parameters
$\beta$ and $\lambda$ are selected by grid search. $\{0.01,0.1,1,10\}$ are tried for both norms in the all cases.
The model with $\beta=1$ and $\lambda=1$ preforms best.


\subsection{Effectiveness of common feature trick} \label{exp_cf}
We prove the effectiveness of common features trick.
Specifically, we set up the experiments with $100$ workers, each of which uses $12$ CPU cores, with up to 110 GB memory totally.
As shown in Table~\ref{comm_fea_cmp}, compressing instances with common feature trick does not affect the actual dimensions of feature space. However, in practice we can significantly reduce memory usage
(reduce to about $1/3$) and speed up the calculation (around 12 times faster) compared to the training
without common feature trick.

\begin{table}
  \centering
  \caption{Training cost comparision with/without common feature trick}
  \begin{tabular}{ccccc}
    \hline
    Cost    &   Without CF.  & With CF. & Cost Saving \\
    \hline
    Memory cost/node&   89.2 GB  &   31 GB & 65.2\%  \\
    \hline
    Time cost/iteration &  121s   &   10s  & 91.7\% \\
    \hline
  \end{tabular}
  \label{comm_fea_cmp}
  \vspace{-5pt}
\end{table}

\begin{figure}[htbp] \label{auc_performance}
    \vspace{-5pt}
    \centering{\includegraphics[width=0.5\textwidth]{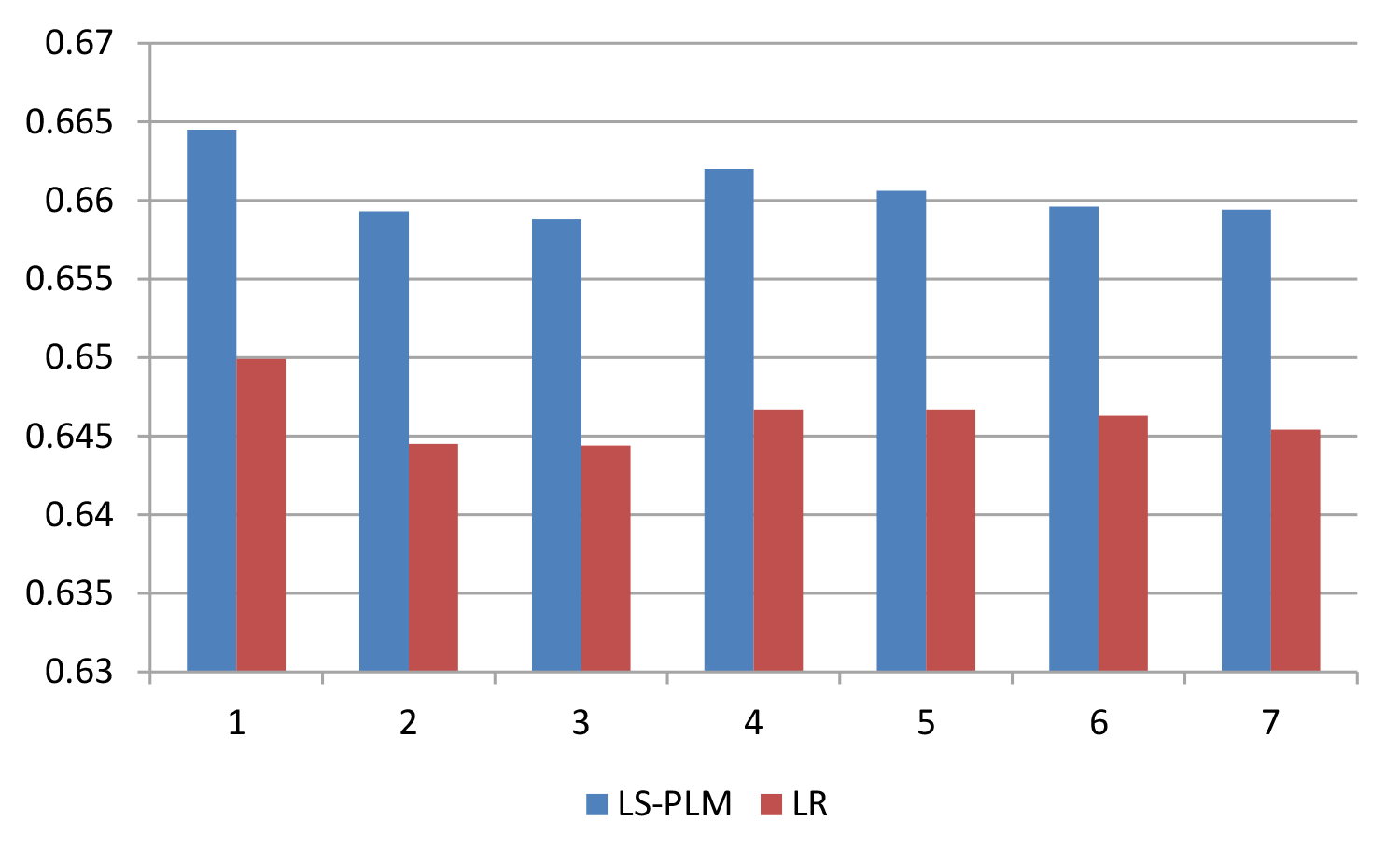}}
    \caption{Model performance comparison on 7 different test datasets. LS-PLM owns consistent and markable promotion compared with LR.}
    \vspace{-5pt}
    \label{auc_performance}
\end{figure}

\subsection{Comparison with LR}
We now compare LS-PLM with LR, the widely used CTR prediction model in product setting.
The two models are trained using our distributed implementation architecture, running with hundreds of machines for speed-up.
The choice of the $L_1$ and $L_{2,1}$ parameters for LS-PLM and the $L_1$ parameter for LR are based on
grid search. $\beta=0.01,0.1,1,10$ and $\lambda=0.01,0.1,1,10$ are tried. The best
parameters are $\beta=1$ and $\lambda=1$ for LS-PLM, and $\beta=1$ for LR.

As shown in Figure \ref{auc_performance}, LS-PLM outperforms LR clearly.
The average improvement of AUC for LR is 1.44\%, which has significant impact to the overall online ad system performance.
Besides, the improvement is stable. This ensures LS-PLM can be safely deployed for daily online production system.


\section{Conclusions}
\label{conclu}

In this paper, a piece-wise linear model, LS-PLM for CTR prediction problem is proposed.
It can capture the  nonlinear pattern from sparse data and save us from heavy feature engineering jobs, which is crucial for real industry applications.
Additionally, powered by our distributed and optimized implementation,
our algorithm can handle problems of billion samples with ten million parameters, which is the typical industrial data volume.
Regularization terms of $L_1$ and $L_{2,1}$ are utilized to keep the model
sparse.
Since 2012, LS-PLM has become the main CTR prediction model in alibaba's online display advertising system, serving hundreds of millions users every day.

\section*{Acknowledgments}
We would like to thank Xingya Dai and Yanghui Yan for their help for this work.


\appendix
\section*{Appendix}
\section{Proof of Lemma 2.1} \label{app_a}

  \begin{proof}

   the definition of $f'(\Theta;d)$ is given as follows:
   \begin{align}
       f'(\Theta;d) =& \lim_{\alpha \downarrow 0} \frac{f(\Theta + \alpha d) - f(\Theta)}{\alpha}  \\ \nonumber
        =& \lim_{\alpha \downarrow 0}\frac{\text{loss}(\Theta + \alpha d) - \text{loss}(\Theta)}{\alpha}   \\ \nonumber
          &+ \lim_{\alpha \downarrow 0} \lambda \frac{\|\Theta + \alpha d\|_{2,1}- \|\Theta\|_{2,1}}{\alpha}     \\ \nonumber
            &+ \lim_{\alpha \downarrow 0} \beta\frac{\|\Theta + \alpha d\|_1 - \|\Theta\|_1}{\alpha}.
   \end{align}
       As the gradient of loss function exists for any $\Theta$, the directional derivative for the first part is
    \begin{equation}
        \lim_{\alpha \downarrow 0}\frac{\text{loss}(\Theta + \alpha d) - \text{loss}(\Theta)}{\alpha} = \nabla \text{loss}(\Theta)^Td \label{dd1}
    \end{equation}
    For the second part, we know if $\|\Theta_{i\cdot}\|_{2,1} \neq 0$, the $L_{2,1}$ norm's partial derivative exists. So the directional derivative is
    \begin{equation}
    \lim_{\alpha \downarrow 0} \lambda \frac{\|\Theta_{i\cdot} + \alpha d_{i\cdot}\|_{2,1}- \|\Theta_{i\cdot}\|_{2,1}}{\alpha} = \lambda \frac{\Theta_{i\cdot}^Td_{i\cdot}}{\|\Theta_{i\cdot}\|_{2,1}}.
    \label{dd2nz}
    \end{equation}
    However, when $\|\Theta_{i\cdot}\|_{2,1} = 0$, it means  $ \Theta_{ij} = 0, 1\leq j \leq 2m$. Then its directional derivative can be denoted as follows:

\begin{align}    \label{dd2z}
  &  \lim_{\alpha \downarrow 0} \lambda \frac{\|\Theta_{i\cdot} + \alpha d_{i\cdot}\|_{2,1}- \|\Theta_{i\cdot}\|_{2,1}}{\alpha}   \\ \nonumber
    &= \lim_{\alpha \downarrow 0}\lambda\frac{\|\alpha d_{i\cdot} \|_{2,1}}{\alpha}  \\ \nonumber
    &= \lambda\|d_{i\cdot}\|_{2,1}
\end{align}

   So combine the above cases in Eq. (\ref{dd2nz}) and Eq. (\ref{dd2z}), we get the directional derivative for the second part:

    \begin{align} \label{dd2}
        & \lim_{\alpha \downarrow 0} \lambda \frac{\|\Theta + \alpha d\|_{2,1}- \|\Theta\|_{2,1}}{\alpha}  \\ \nonumber
        & = \sum_{\|\Theta_{i\cdot}\|_{2,1} \neq 0} \lambda \frac{\Theta_{i\cdot}^Td_{i\cdot}}{\|\Theta_{i\cdot}\|_{2,1}} + \sum_{\|\Theta_{i\cdot}\|_{2,1} =0} \lambda \|d_{i\cdot}\|_{2,1}
    \end{align}

    Same as the second part, the direction derivative for the third part is:
 \begin{align} \label{dd3}
   & \lim_{\alpha \downarrow 0} \beta \frac{\|\Theta + \alpha d\|_1- \|\Theta\|_1}{\alpha}  \\ \nonumber
   &= \sum_{ \|\Theta_{ij}\|_1 \neq 0} \beta \text{sign}(\Theta_{ij})d_{ij} + \sum_{\|\Theta_{ij}\|_1 =0} \beta |d_{ij}|
 \end{align}

   Based on Eq. (\ref{dd1}), Eq. (\ref{dd2}) and Eq. (\ref{dd3}), we get that for any $\Theta$ and direction $d$, $f'(\Theta;d)$ exists.

  \end{proof}

\section{Proof of Proposition 2.2} \label{app_b}
\begin{proof}

Finding the expected direction turns to an optimization problem, which is formulated as follows:
\begin{equation}
   \min_d f'(\Theta;d) \quad \text{s.t. } \|d\|^2 \leq C.
\end{equation}
Here the direction $d$ is bounded by a constant scalar $C$. To solve this problem, we employ lagrange function to combine the objective function and inequality
function together:
\begin{equation}
     L(d,\mu) = f'(\Theta;d) + \mu(\|d\|^2 - C).
\end{equation}
Here $\mu \geq 0$ is the lagrange multiplier. Setting the partial derivative of $d$ with respect to $L(d,\mu)$ to zero has three cases.

Define $ s = -\nabla loss(\Theta)_{ij}- \lambda\frac{\Theta_{ij}}{\|\Theta_{i\cdot}\|_{2,1}} $

a. when $\Theta_{ij} \neq 0$, it implies that
   \begin{equation}
   2\mu d_{ij} = s- \beta \text{sign}(\Theta_{ij}) \nonumber
   \end{equation}

b. when $\Theta_{ij} = 0$ and $\|\Theta_{i\cdot}\|_{2,1} >  0$, it is easy to have
\begin{equation}
2\mu d_{ij} =  \max\{ |s|- \beta, 0 \} \text{sign}(s) \nonumber
\end{equation}

c. We give more details when $\Theta_{ij} = 0$ and $\|\Theta_{i\cdot}\|_{2,1} = 0$. For $d_{i\cdot}$ we have
\begin{equation}
   \frac{\partial L(d,\mu)}{\partial d_{i\cdot}} = \nabla \text{loss}(\Theta)_{i\cdot} + \beta \text{sign}(d_{i\cdot}) + \lambda \frac{d_{i\cdot}}{\|d_{i\cdot}\|_{2,1}} + 2\mu d_{i\cdot}=0.
   \nonumber
\end{equation}
Here we use $\text{sign}(d_{i\cdot}) = [  \text{sign}(d_{i1}),\dots, \text{sign}(d_{i 2m}) ]^T $ for simplicity. Then we get
\begin{equation}
  (2\mu + \frac{\lambda}{\|d_{i\cdot}\|_{2,1}}) d_{i\cdot} = - \nabla \text{loss}(\Theta)_{i\cdot} - \beta \text{sign}(d_{i\cdot})
  \nonumber
\end{equation}
which implies that $\text{sign}(d_{i\cdot}) =  \text{sign}(- \nabla loss(\Theta)_{i\cdot} - \beta \text{sign}(d_{i\cdot})) $. When $d_{ij} \geq 0 $, it implies $- \nabla \text{loss}(\Theta)_{ij} - \beta \text{sign}(d_{ij}) \geq 0$. Inversely, we have  $- \nabla \text{loss}(\Theta)_{ij} - \beta \text{sign}(d_{ij}) \leq 0$ when $d_{ij} \leq 0.$ So we define $v \doteq- \nabla \text{loss}(\Theta)_{i\cdot} - \beta \text{sign}(d_{i\cdot})$ and $v_j  =  \max\{|-\nabla \text{loss}(\Theta)_{ij}| - \beta,0\}\text{sign}(-\nabla \text{loss}(\Theta)_{ij})$. So
\begin{eqnarray}
 &(2\mu + \frac{\lambda}{\|d_{i\cdot}\|_{2,1}}) d_{i\cdot} = v \nonumber \\
\Rightarrow &(2u\|d_{i\cdot}\|  +  \lambda) \|d_{i\cdot}\| = \|v\| \|d_{i\cdot}\|  \nonumber \\
\Rightarrow & (2u\|d_{i\cdot}\|  +  \lambda) = \|v\|. \nonumber
\end{eqnarray}

As $\|d_{i\cdot}\| \geq 0$, we have $2\mu \|d_{i\cdot}\| = \max(\|v\| - \lambda,0) $. Thus $2\mu d_{ij} = \frac{\max(\|v\| - \lambda,0) }{\|v\|} v$

The lagrange multiplier $u$ is a scalar, and it has equivalent influence for all $d_{ij}$. We can see that the optimal direction which is bounded by $C$ is the same direction as we defined in Eq. (\ref{mlrdir}) without considering the constant scalar $\mu$. Here we finish our proof.

\end{proof}

%
%

\begin{thebibliography}{10}

\bibitem{blei2003latent}
David~M Blei, Andrew~Y Ng, and Michael~I Jordan.
\newblock Latent dirichlet allocation.
\newblock {\em the Journal of machine Learning research}, 3:993--1022, 2003.

\bibitem{donahue2014long}
Jeff Donahue, Lisa~Anne Hendricks, Sergio Guadarrama, Marcus Rohrbach,
  Subhashini Venugopalan, Kate Saenko, and Trevor Darrell.
\newblock Long-term recurrent convolutional networks for visual recognition and
  description.
\newblock {\em arXiv}, 2014.

\bibitem{karpathy2014deep}
Andrej Karpathy and Li~Fei-Fei.
\newblock Deep visual-semantic alignments for generating image descriptions.
\newblock {\em arXi}, 2014.

\bibitem{kingma2014adam}
Diederik Kingma and Jimmy Ba.
\newblock Adam: A method for stochastic optimization.
\newblock {\em arXiv}, 2014.

\bibitem{kiros2014unifying}
Ryan Kiros, Ruslan Salakhutdinov, and Richard~S Zemel.
\newblock Unifying visual-semantic embeddings with multimodal neural language
  models.
\newblock {\em arXiv}, 2014.

\bibitem{krizhevsky2012imagenet}
Alex Krizhevsky, Ilya Sutskever, and Geoffrey~E Hinton.
\newblock Imagenet classification with deep convolutional neural networks.
\newblock In {\em NIPS}, 2012.

\bibitem{manning2014stanford}
Christopher~D Manning, Mihai Surdeanu, John Bauer, Jenny Finkel, Steven~J
  Bethard, and David McClosky.
\newblock The stanford corenlp natural language processing toolkit.
\newblock In {\em ACL}, 2014.

\bibitem{mao2014explain}
Junhua Mao, Wei Xu, Yi~Yang, Jiang Wang, and Alan~L Yuille.
\newblock Explain images with multimodal recurrent neural networks.
\newblock {\em arXiv}, 2014.

\bibitem{vinyals2014show}
Oriol Vinyals, Alexander Toshev, Samy Bengio, and Dumitru Erhan.
\newblock Show and tell: A neural image caption generator.
\newblock {\em arXiv}, 2014.

\bibitem{xu2015show}
Kelvin Xu, Jimmy Ba, Ryan Kiros, Aaron Courville, Ruslan Salakhutdinov, Richard
  Zemel, and Yoshua Bengio.
\newblock Show, attend and tell: Neural image caption generation with visual
  attention.
\newblock {\em arXiv}, 2015.

\bibitem{zhou2014places}
B.~Zhou, A.~Lapedriza, J.~Xiao, A.~Torralba, and A.~Oliva.
\newblock {Learning Deep Features for Scene Recognition using Places Database.}
\newblock {\em NIPS}, 2014.

\end{thebibliography}


\begin{thebibliography}{10}

\bibitem{bahdanau2014neural}
Dzmitry Bahdanau, Kyunghyun Cho, and Yoshua Bengio.
\newblock Neural machine translation by jointly learning to align and
  translate.
\newblock {\em arXiv}, 2014.

\bibitem{blei2003latent}
David~M Blei, Andrew~Y Ng, and Michael~I Jordan.
\newblock Latent dirichlet allocation.
\newblock {\em the Journal of machine Learning research}, 3:993--1022, 2003.

\bibitem{chen2015microsoft}
Xinlei Chen, Hao Fang, Tsung-Yi Lin, Ramakrishna Vedantam, Saurabh Gupta, Piotr
  Dollar, and C~Lawrence Zitnick.
\newblock Microsoft coco captions: Data collection and evaluation server.
\newblock {\em arXiv}, 2015.

\bibitem{donahue2014long}
Jeff Donahue, Lisa~Anne Hendricks, Sergio Guadarrama, Marcus Rohrbach,
  Subhashini Venugopalan, Kate Saenko, and Trevor Darrell.
\newblock Long-term recurrent convolutional networks for visual recognition and
  description.
\newblock {\em arXiv}, 2014.

\bibitem{elliott2013image}
Desmond Elliott and Frank Keller.
\newblock Image description using visual dependency representations.
\newblock In {\em EMNLP}, 2013.

\bibitem{fang2014captions}
Hao Fang, Saurabh Gupta, Forrest Iandola, Rupesh Srivastava, Li~Deng, Piotr
  Doll{\'a}r, Jianfeng Gao, Xiaodong He, Margaret Mitchell, John Platt, et~al.
\newblock From captions to visual concepts and back.
\newblock {\em arXiv}, 2014.

\bibitem{hochreiter1997long}
Sepp Hochreiter and J{\"u}rgen Schmidhuber.
\newblock Long short-term memory.
\newblock {\em Neural computation}, 9(8):1735--1780, 1997.

\bibitem{karpathy2014deep}
Andrej Karpathy and Li~Fei-Fei.
\newblock Deep visual-semantic alignments for generating image descriptions.
\newblock {\em arXi}, 2014.

\bibitem{kiros2014unifying}
Ryan Kiros, Ruslan Salakhutdinov, and Richard~S Zemel.
\newblock Unifying visual-semantic embeddings with multimodal neural language
  models.
\newblock {\em arXiv}, 2014.

\bibitem{krizhevsky2012imagenet}
Alex Krizhevsky, Ilya Sutskever, and Geoffrey~E Hinton.
\newblock Imagenet classification with deep convolutional neural networks.
\newblock In {\em NIPS}, 2012.

\bibitem{kulkarni2013babytalk}
Girish Kulkarni, Visruth Premraj, Vicente Ordonez, Sagnik Dhar, Siming Li,
  Yejin Choi, Alexander~C Berg, and Tamara~L Berg.
\newblock Babytalk: Understanding and generating simple image descriptions.
\newblock {\em Pattern Analysis and Machine Intelligence, IEEE Transactions
  on}, 35(12):2891--2903, 2013.

\bibitem{kuznetsova2012collective}
Polina Kuznetsova, Vicente Ordonez, Alexander~C Berg, Tamara~L Berg, and Yejin
  Choi.
\newblock Collective generation of natural image descriptions.
\newblock In {\em ACL}, 2012.

\bibitem{kuznetsova2014treetalk}
Polina Kuznetsova, Vicente Ordonez, Tamara~L Berg, and Yejin Choi.
\newblock Treetalk: Composition and compression of trees for image
  descriptions.
\newblock {\em Transactions of the Association for Computational Linguistics},
  2(10):351--362, 2014.

\bibitem{li2011composing}
Siming Li, Girish Kulkarni, Tamara~L Berg, Alexander~C Berg, and Yejin Choi.
\newblock Composing simple image descriptions using web-scale n-grams.
\newblock In {\em CoNLL}, 2011.

\bibitem{lin2014microsoft}
Tsung-Yi Lin, Michael Maire, Serge Belongie, James Hays, Pietro Perona, Deva
  Ramanan, Piotr Doll{\'a}r, and C~Lawrence Zitnick.
\newblock Microsoft coco: Common objects in context.
\newblock In {\em ECCV}. 2014.

\bibitem{mao2014explain}
Junhua Mao, Wei Xu, Yi~Yang, Jiang Wang, and Alan~L Yuille.
\newblock Explain images with multimodal recurrent neural networks.
\newblock {\em arXiv}, 2014.

\bibitem{mitchell2012midge}
Margaret Mitchell, Xufeng Han, Jesse Dodge, Alyssa Mensch, Amit Goyal, Alex
  Berg, Kota Yamaguchi, Tamara Berg, Karl Stratos, and Hal Daum{\'e}~III.
\newblock Midge: Generating image descriptions from computer vision detections.
\newblock In {\em EACL}, 2012.

\bibitem{rashtchian2010collecting}
Cyrus Rashtchian, Peter Young, Micah Hodosh, and Julia Hockenmaier.
\newblock Collecting image annotations using amazon's mechanical turk.
\newblock In {\em NAACL workshop}, 2010.

\bibitem{Simonyan14c}
K.~Simonyan and A.~Zisserman.
\newblock Very deep convolutional networks for large-scale image recognition.
\newblock {\em CoRR}, abs/1409.1556, 2014.

\bibitem{simonyan2014very}
Karen Simonyan and Andrew Zisserman.
\newblock Very deep convolutional networks for large-scale image recognition.
\newblock {\em arXiv}, 2014.

\bibitem{sutskever2011generating}
Ilya Sutskever, James Martens, and Geoffrey~E Hinton.
\newblock Generating text with recurrent neural networks.
\newblock In {\em ICML}, 2011.

\bibitem{taylor2009factored}
Graham~W Taylor and Geoffrey~E Hinton.
\newblock Factored conditional restricted boltzmann machines for modeling
  motion style.
\newblock In {\em ICML}, 2009.

\bibitem{uijlings2013selective}
Jasper~RR Uijlings, Koen~EA van~de Sande, Theo Gevers, and Arnold~WM Smeulders.
\newblock Selective search for object recognition.
\newblock {\em International journal of computer vision}, 104(2):154--171,
  2013.

\bibitem{vinyals2014show}
Oriol Vinyals, Alexander Toshev, Samy Bengio, and Dumitru Erhan.
\newblock Show and tell: A neural image caption generator.
\newblock {\em arXiv}, 2014.

\bibitem{xu2015show}
Kelvin Xu, Jimmy Ba, Ryan Kiros, Aaron Courville, Ruslan Salakhutdinov, Richard
  Zemel, and Yoshua Bengio.
\newblock Show, attend and tell: Neural image caption generation with visual
  attention.
\newblock {\em arXiv}, 2015.

\bibitem{yang2011corpus}
Yezhou Yang, Ching~Lik Teo, Hal Daum{\'e}~III, and Yiannis Aloimonos.
\newblock Corpus-guided sentence generation of natural images.
\newblock In {\em EMNLP}, 2011.

\bibitem{young2014image}
Peter Young, Alice Lai, Micah Hodosh, and Julia Hockenmaier.
\newblock From image descriptions to visual denotations: New similarity metrics
  for semantic inference over event descriptions.
\newblock {\em Transactions of the Association for Computational Linguistics},
  2:67--78, 2014.

\end{thebibliography}


\begin{thebibliography} {400}
\large{
\bibitem{AndrewICML}
Andrew G. and Gao J. (2007) Scalable Training of $L_{1}$-Regularized Log-Linear Models.
\emph{Proceedings of the 24-th International Conference on Machine Learning}.

\bibitem{LBFGS}
Bertsekas, D. (2003) \emph{Nonlinear Programming}. Springer US, 51--88.

\bibitem{McMahanGoogle}
Brendan H., Holt G., Sculley D., Young M.,
Ebner D., Grady J., Nie L., Phillips. T, Davydov E.,
Golovin D., Chikkerur S., Liu D., Wattenberg M.,
Hrafnkelsson A., Boulos T., Kubica J. (2013)
Ad Click Prediction: a View from the Trenches. \emph{Proceedings of the 19-th KDD}.

\bibitem{FawcettAUC}
Fawcett T. (2006) An introduction to ROC analysis. \emph{Pattern Recognition Letters}, 27, 861--874.

\bibitem{FriedmanGBDT}
Friedman J. (1999) Greedy Function Approximation: A Gradient Boosting Machine.
\emph{Technical Report, Dept. of Statistics}, Stanford University.

\bibitem{LRModel}
Hilbe M. (2009) Logistic regression models. CRC Press.

\bibitem{Facebook-GBDT+LR}
He X., Pan J., Jin O., Xu T., Liu B., Xu T, Shi Y.,
Atallah A, Herbrich R., Bowers S., Candela J. (2014)
Practical Lessons from Predicting Clicks on Ads at Facebook.
\emph{Proceedings of the 20-th KDD}.

\bibitem{JORDANFOE}
Jordan I., Jacobs A (1994) Hierarchical mixtures of experts and the EM algorithm. \emph{Neural computation}, 6(2): 181-214.

\bibitem{Softmax}
Kivinen J., Warmuth M K. (1998) Relative Loss Bounds for Multidimensional Regression Problems. \emph{Machine Learning}, 45(3):301-329.
%
\bibitem{RendleFM}
Rendle S. (2010) Factorization Machines. \emph{Proceedings of the 10th IEEE International
Conference on Data Mining}.



\bibitem{FOE}
Roth S, Black M J. (2009) Fields of experts. \emph{International Journal of Computer Vision}, 82(2): 205--229.

\bibitem{DTREE}
Safavian S. R., Landgrebe D. (1990) A survey of decision tree classifier methodology[J].

\bibitem{WangMLR}
Wang P.-M and Puterman M. (1998) Mixed Logistic Regression Models.
\emph{Journal of Agricultural, Biological, and Environmental Statistics}, 3(2), 175--200.


\bibitem{ZHANGSGD}
Zhang T. (2004) Solving large scale linear prediction problems using stochastic gradient
descent algorithms. \emph{Proceedings of the twenty-first international conference on Machine learning}. ACM, 116.

\bibitem{url}
Gai K. \url{http://club.alibabatech.org/resource_detail.htm?topicId=106}

}
\end{thebibliography}
%

\end{document}